\documentclass[10pt,twocolumn,letterpaper]{article}

\usepackage[pagenumbers]{cvpr} % To force page numbers, e.g. for an arXiv version

\usepackage{graphicx}
\usepackage{amsmath}
\usepackage{amssymb}
\usepackage{booktabs}
\usepackage{xcolor}

\usepackage[pagebackref,breaklinks,colorlinks]{hyperref}

\usepackage[capitalize]{cleveref}
\crefname{section}{Sec.}{Secs.}
\Crefname{section}{Section}{Sections}
\Crefname{table}{Table}{Tables}
\crefname{table}{Tab.}{Tabs.}

\def\cvprPaperID{888} % *** Enter the CVPR Paper ID here
\def\confName{CVPR}
\def\confYear{2022}

\begin{document}

\title{A Competitive Method for Dog Nose-print Re-identification}

\author{Fei Shen\textsuperscript{1}, Zhe Wang\textsuperscript{2}, Zijun Wang\textsuperscript{3}, Xiaode Fu\textsuperscript{1}, Jiayi Chen\textsuperscript{1}, Xiaoyu Du\textsuperscript{1} and Jinhui Tang\textsuperscript{1}\\
\\
\textsuperscript{1} Nanjing University of Science and Technology\\
\textsuperscript{2} DeepBlue Technology Co., Ltd\\
\textsuperscript{3} Guangdong University of Technology\\
{\tt\small feishen@njust.edu.cn}

}

\maketitle

%%%%%%%%% ABSTRACT
\begin{abstract}
Vision-based pattern identification (such as face, fingerprint, iris etc.) has been successfully applied in human biometrics for a long history. However, dog nose-print authentication is a challenging problem since the lack of a large amount of labeled data. For that, this paper presents our proposed methods for dog nose-print authentication (Re-ID) task in CVPR 2022 pet biometric challenge. First, considering the problem that each class only with few samples in the training set, we propose an automatic offline data augmentation strategy. Then, for the difference in sample styles between the training and test datasets, we employ joint cross-entropy, triplet and pair-wise circle losses function for network optimization. Finally, with multiple models ensembled adopted, our methods achieve 86.67\% AUC on the test set. Codes are available at \url{https://github.com/muzishen/Pet-ReID-IMAG.}
\end{abstract}

%%%%%%%%% BODY TEXT
\section{Introduction}
\label{sec:intro}

More and more families choose to keep some pets to accompany them in recent years. According to the GMI report, global pet care market size surpassed 232 billion in 2020. With the rapid growth of pet economy, pet identification is a challenging problem in many scenarios such as pet management, trading, insurance, medical treatment etc., unfortunately, there is no solution balanced accuracy, cost and usability well for this challenge up to now.

\begin{figure}[tp]
	\centering
	\includegraphics[width=1.0\linewidth]{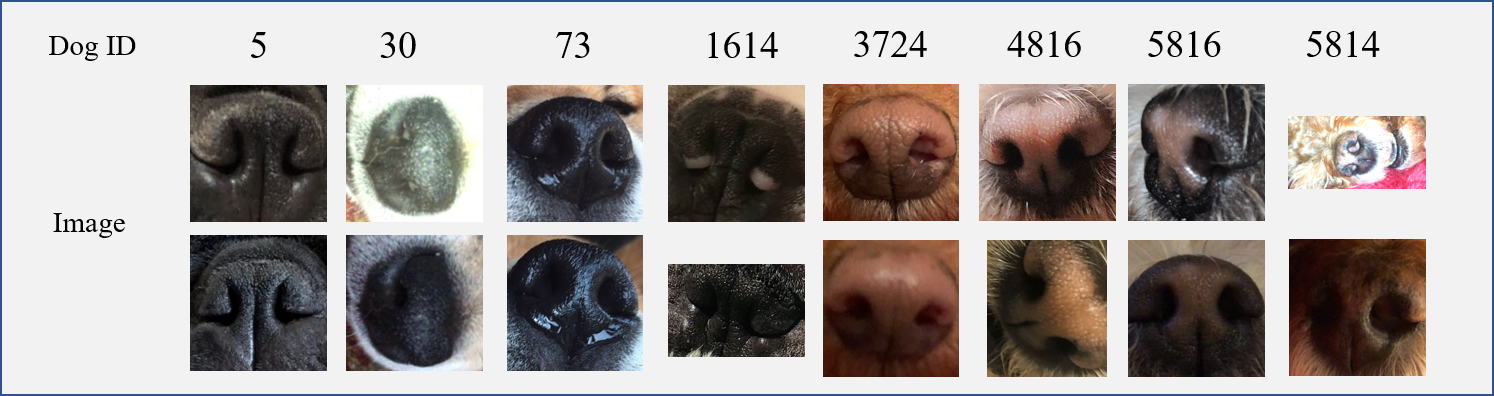}
    \vspace{-0.3cm}
	\caption{The example of training data. Each column represents the same ID.}\label{fig:example}
\end{figure}

In human biometrics, person/vehicle re-identification (Re-ID) \cite{hpgn, luobot, alignedreid, git, emrn, transreid, hsgm, li2021triple} methods based on deep learning have made a significant process in recent years.
Pet biometric challenge \footnote{https://www.vislab.ucr.edu/Biometrics2022/index.php} is a workshop in the ECCV2020 conference.
The challenge focuses on obtaining high area under curve (AUC) on a dog nose-print dataset.
It is very challenging for dog nose-print re-identification due to the adverse influence of the sample class imbalance and lacking of labeled data, as shown in \ref{fig:example}.
However, we find that 1 vs 1 pet identity verification by dog nose-print images is very similar to the pedestrian Re-ID task.
The two tasks all need to train a model to extract features for each identity, and then compare the extracted features to judge id information. Based on the pipeline of pedestrian Re-ID methods, we designed the framework of 1 vs 1 pet identity verification.

	The rest of the paper is organized as follows. In Section  \ref{sec:method}, the proposed methods is introduced. The experimental results are presented in Section \ref{sec:exp}. And finally Section \ref{sec:con} concludes the paper.

\section{Methods}\label{sec:method}
\begin{figure}[tp]
	\centering
	\includegraphics[width=.9\linewidth]{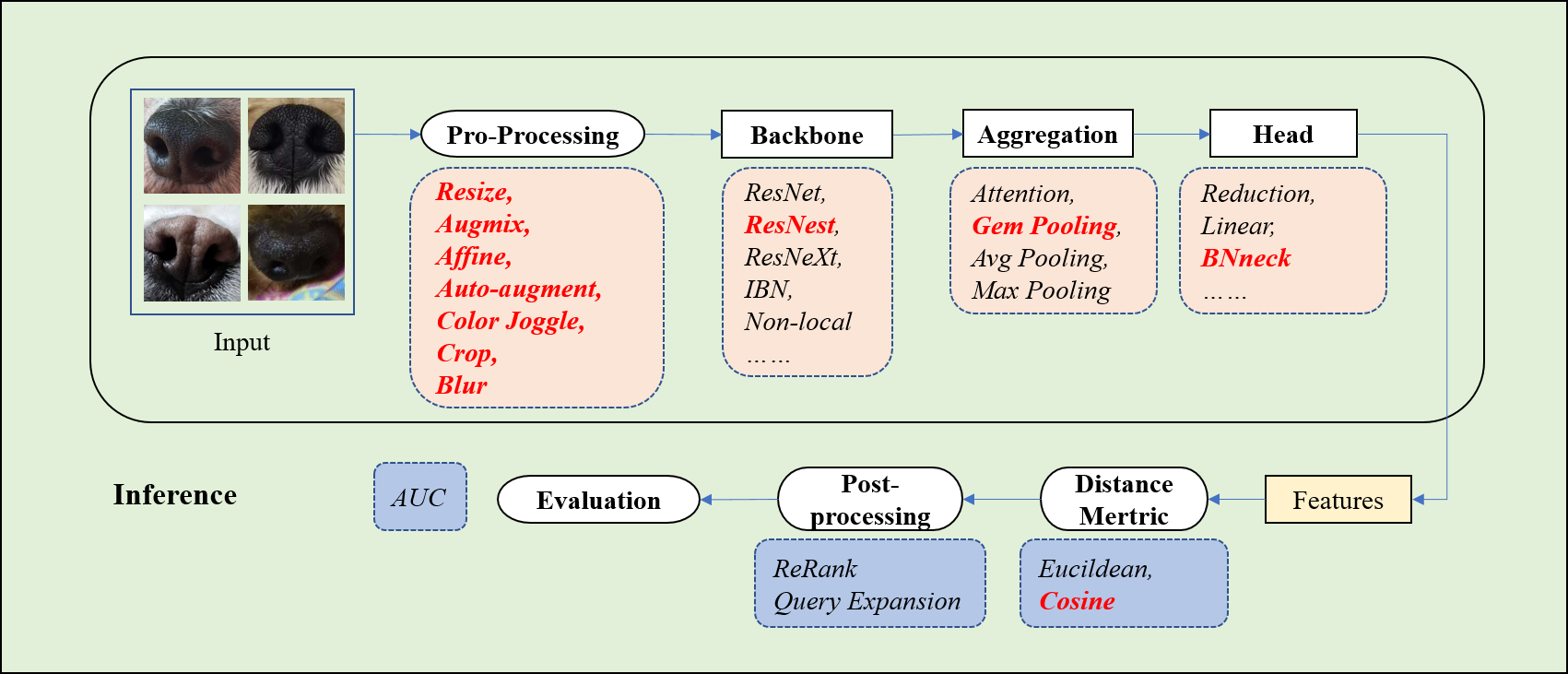}
    \vspace{-0.3cm}
	\caption{The pipeline for pet ReID.}\label{fig:pipline}

\end{figure}

The pipeline of our proposed method is shown in Figure \ref{fig:pipline} and consists of input image pro-processing module, backbone, aggregation module and head. We will introduce them in detail in the follow.

\subsection{Image Pro-processing}
	The training images are of different sizes, we first resize the image to the fixed-size images so that input images can be collected into batches and input into the backbone. To obtain a more robust model, affine and crop as data augmentation methods by applying affine transformation and crop operation to image to make the model better adapted to shape and size changes.
In addition, performing color joggle on the image makes data more diverse. Considering the data shift problem between the training set and test set, we perform the AugMix augmentation method on the training set, which randomly apply different data augmentations to the image (Aug) and then mix multiple images (Mix). Auto-augment is based on automl technique to achieve effective data augmentation for improving the robustness of feature representation.
Blur is also a important data augmentation method to extract important information from images.
\subsection{Backbone}

	Backbone is the important module for image feature extraction. In our pet-ReID pipeline, we use three different backbone (ResNet \cite{resnet}, ResNeSt \cite{resnest}, ResNeXt \cite{resnext}) to get image feature maps. ResNet is a widely used backbone network for CV tasks, and the residual module in ResNet allows the network to go deeper and thus learn better features. ResNeSt explores a simple architectural modification of the ResNet, incorporating feature-map split attention within the individual network blocks which allows attention operation across the feature-map groups. ResNeXt adopts VGG/ResNets’ strategy of repeating layers, while exploiting the split-transform-merge strategy in an easy extensible way. We also add attention-like non-local \cite{nonlocal} module  and instance batch normalization (IBN) \cite{ibn} module into backbones to learn more robust feature.

\begin{table}[tp]
	\centering
	\caption{
		 Comparison results with ResNeSt101 on validation set.
	}\label{tab:1}
	\begin{tabular}{l|c}
                {Methods}                   &AUC  \\

        \hline

        Baseline					&87.2 		\\
		+ Augmix					&88.4		\\
		+ Affine					&88.9 		 \\
		+ Auto-augment              &90.1   \\
		+ Color jitter       		&90.3       \\	
		+ Blur                       & 91.5   \\
		+ Crop                       & 91.7  \\
\hline
	\end{tabular}
\end{table}

\begin{table}[tp]
	\centering
	\caption{
		 Comparison results with different backbone on validation set.
	}\label{tab:2}
	\begin{tabular}{l|c}
                {Methods}                   &AUC  \\

        \hline

        ResNet-ibn-101 \cite{ibn}					&88.2 		\\
		ResNeXt-101	 \cite{resnext}				&90.3		\\
		Swin-transformer-base \cite{swin}					&89.8 		 \\
		ConvNet-base    \cite{convnet}          &90.3   \\
		Res2Net-101 \cite{res2net}      		    &90.5       \\	
		ResNeSt-101  \cite{resnest}                     & 91.7   \\
		ResNeSt-200      \cite{resnest}                 & 91.9  \\
\hline
	\end{tabular}
\end{table}

\begin{table}[tp]
	\centering
	\caption{
		 The performance on testing set with single model and model ensemble.
	}\label{tab:3}
	\begin{tabular}{l|c}
                {Methods}                   &AUC  \\

        \hline

        Single Model					&86.1		\\
		Model Ensemble					&86.7		\\

\hline
	\end{tabular}
\end{table}

\subsection{Aggregation}
	The aggregation module aims to aggregate image feature maps generated by the backbone into a global feature. In our work, we apply four aggregation methods to the feature map, namely attention pooling, GeM pooling, average pooling and max pooling.
Head
	Head is the part of addressing the global vector generated by aggregation module. In our pet-ReID works, we use three different head, including batch normalization (BN) head, Linear head and Reduction head. Each head contains one or more of BN layer, reduction layer and decision layer. The linear head only contains a decision layer, the BN head contains a BN layer and a decision layer and the reduction head contains conv+bn+relu+dropout operation, a reduction layer and a decision layer, where batch normalization is used to solve internal covariate shift because it is very difficult to train models with saturating non-linearities, reduction layer is aiming to make the high-dimensional feature become the low-dimensional feature and decision layer outputs the probability of different IDs to distinguish different IDs for the following model training.

\subsection{Testing}
	In the test phase, the test image is input into the model to get feature representation. Then the extracted feature is compared with the features in the feature library for the distance metric such as Eucildean and cosine measure. Thereafter, the results are post-processed by Query Expansion which is a re-rank method. The flow of Query Expansion is as follows: Given a query image, and use it to find m similar gallery images. The query feature is defined as $f_q$ and m similar gallery features are defined as $f_g$. Then the new query feature is constructed by averaging the verified gallery features and the query feature.

\section{Experiments} \label{sec:exp}
The model structure is based on Fast-ReID \cite{fastreid}. We trained the models (i.e., ResNeSt \cite{resnest}, ResNet-ibn \cite{ibn}, ResNeXt \cite{resnext}, swin-transformer \cite{swin}, ConvNet \cite{convnet}, Res2Net \cite{res2net}) on different backbones pretrained on ImageNet.
Label-smoothed cross entropy loss is adopted for classification.
The soft-margin triplet and circle losses are adopted for metric learning.

\subsection{Implement Detail}
All experiments are conducted using the Fast-reid \cite{fastreid} toolbox developed by PyTorch.
And we run experiments on a NVIDIA V100 GPU with 16GB.
Training configurations are summarized as follows \cite{hpgn}.
(1)The input images are randomly sized to $224 \times 224$, $256 \times 256$, and $288 \times 288$.
(2)For both cutmix and random flip operations, the implementation probability is set to 0.5.
(3)The mini-batch Adam method  is applied to optimize parameters. The weight decays are set to 5$\times$10$^{-4}$, and the momentums are set to 0.9.
(4)Each mini-batch includes 64 vehicle images, which includes 16 subjects and each subject holds 4 images.
(5) The initial learning is fixed to 0.00035.  The model is trained for 35 epochs in total.

\subsection{Ablation Study}
In this section, we design different ablation studies to evaluate the effectiveness of our method.

In our work, the proposed method uses several techniques on the baseline to improve the recognition accuracy, e.g. augmix, affine, and model ensemble. We show the ablation study of our image pre-processing technique in Table \ref{tab:1}. For example, a model with augmix is proven to work due to 1.2\% higher than the baseline. After applying all these techniques, our accuracy increases by 4.5\%, from 87.2\% to 91.7\% on the validation test.

Also, we test several backbones with our pre-processing techniques. As the experimental results using different backbones on our validation set are shown in Table \ref{tab:2}, ResNeSt stands out from all backbones with over 91.7\% accuracy.

In Table \ref{tab:3}, we use a model ensemble. We combine ResNeSt-101's 224, 256, and 288 scale feature maps with 224 map of ResNeSt-200 for multi-scale fusion on testing set. We achieve a 0.6\% growth compared with the single model, which proves the effectiveness of model ensemble.
Finally, our method get an 86.7\% score in the competition.

\section{Conclusions} \label{sec:con}
This report details the key technologies used in the Pet Biometric challenge.
 Our primary concern is the data augmentation to extract more compelling features.
 The introduction of Aug-mix, Affine, Auto-augmentation, color-jitter, blur and crop to expand the training set make the model more robust.
  Extensive experiments on a subset of the  dataset of dog nose-print demonstrate that the proposed method can obtain a competitive performance, which
can be further analyzed and better utilized in future works.
%%%%%%%% REFERENCES
{\small
\bibliographystyle{ieee_fullname}
\bibliography{egbib}

\begin{thebibliography}{10}\itemsep=-1pt

\bibitem{res2net}
Shang-Hua Gao, Ming-Ming Cheng, Kai Zhao, Xin-Yu Zhang, Ming-Hsuan Yang, and
  Philip Torr.
\newblock Res2net: A new multi-scale backbone architecture.
\newblock {\em IEEE TPAMI}, 2020.

\bibitem{resnet}
Kaiming He, Xiangyu Zhang, Shaoqing Ren, and Jian Sun.
\newblock Deep residual learning for image recognition.
\newblock In {\em Conference on Computer Vision and Pattern Recognition}, pages
  770--778, 2016.

\bibitem{fastreid}
Lingxiao He, Xingyu Liao, Wu Liu, Xinchen Liu, Peng Cheng, and Tao Mei.
\newblock Fastreid: A pytorch toolbox for general instance re-identification.
\newblock {\em arXiv preprint arXiv:2006.02631}, 2020.

\bibitem{transreid}
Shuting He, Hao Luo, Pichao Wang, Fan Wang, Hao Li, and Wei Jiang.
\newblock Transreid: Transformer-based object re-identification.
\newblock In {\em Proceedings of the IEEE/CVF International Conference on
  Computer Vision}, pages 15013--15022, 2021.

\bibitem{li2021triple}
Huafeng Li, Neng Dong, Zhengtao Yu, Dapeng Tao, and Guanqiu Qi.
\newblock Triple adversarial learning and multi-view imaginative reasoning for
  unsupervised domain adaptation person re-identification.
\newblock {\em IEEE Transactions on Circuits and Systems for Video Technology},
  32(5):2814--2830, 2021.

\bibitem{swin}
Ze Liu, Yutong Lin, Yue Cao, Han Hu, Yixuan Wei, Zheng Zhang, Stephen Lin, and
  Baining Guo.
\newblock Swin transformer: Hierarchical vision transformer using shifted
  windows.
\newblock In {\em Proceedings of the IEEE/CVF International Conference on
  Computer Vision}, pages 10012--10022, 2021.

\bibitem{convnet}
Zhuang Liu, Hanzi Mao, Chao-Yuan Wu, Christoph Feichtenhofer, Trevor Darrell,
  and Saining Xie.
\newblock A convnet for the 2020s.
\newblock {\em arXiv preprint arXiv:2201.03545}, 2022.

\bibitem{luobot}
H. {Luo}, W. {Jiang}, Y. {Gu}, F. {Liu}, X. {Liao}, S. {Lai}, and J. {Gu}.
\newblock A strong baseline and batch normalization neck for deep person
  re-identification.
\newblock {\em IEEE Transactions on Multimedia}, pages 1--1, 2019.

\bibitem{alignedreid}
Hao Luo, Wei Jiang, Xuan Zhang, Xing Fan, Jingjing Qian, and Chi Zhang.
\newblock Alignedreid++: Dynamically matching local information for person
  re-identification.
\newblock {\em Pattern Recognition}, 94:53--61, 2019.

\bibitem{ibn}
Xingang Pan, Ping Luo, Jianping Shi, and Xiaoou Tang.
\newblock Two at once: Enhancing learning and generalization capacities via
  ibn-net.
\newblock In {\em European Conference on Computer Vision}, pages 464--479,
  2018.

\bibitem{hsgm}
Fei Shen, Xiaoxiao Peng, Lisheng Wang, Xingmeng Zhang, Mei Shu, and Yayun Wang.
\newblock Hsgm: A hierarchical similarity graph module for object
  re-identification.
\newblock In {\em 2022 IEEE International Conference on Multimedia and Expo
  (ICME)}, pages 1--6. IEEE, 2022.

\bibitem{git}
Fei Shen, Yi Xie, Jianqing Zhu, Xiaobin Zhu, and Huanqiang Zeng.
\newblock Git: Graph interactive transformer for vehicle re-identification.
\newblock {\em arXiv preprint arXiv:2107.05475}, 2021.

\bibitem{emrn}
Fei Shen, Jianqing Zhu, Xiaobin Zhu, Jingchang Huang, Huanqiang Zeng, Zhen Lei,
  and Canhui Cai.
\newblock An efficient multi-resolution network for vehicle re-identification.
\newblock {\em IEEE Internet of Things Journal}, 2021.

\bibitem{hpgn}
Fei Shen, Jianqing Zhu, Xiaobin Zhu, Yi Xie, and Jingchang Huang.
\newblock Exploring spatial significance via hybrid pyramidal graph network for
  vehicle re-identification.
\newblock {\em IEEE Transactions on Intelligent Transportation Systems}, 2021.

\bibitem{nonlocal}
Xiaolong Wang, Ross Girshick, Abhinav Gupta, and Kaiming He.
\newblock Non-local neural networks.
\newblock In {\em Proceedings of the IEEE conference on computer vision and
  pattern recognition}, pages 7794--7803, 2018.

\bibitem{resnext}
Saining Xie, Ross Girshick, Piotr Dollár, Zhuowen Tu, and Kaiming He.
\newblock Aggregated residual transformations for deep neural networks.
\newblock {\em arXiv preprint arXiv:1611.05431}, 2016.

\bibitem{resnest}
Hang Zhang, Chongruo Wu, Zhongyue Zhang, Yi Zhu, Haibin Lin, Zhi Zhang, Yue
  Sun, Tong He, Jonas Mueller, R Manmatha, et~al.
\newblock Resnest: Split-attention networks.
\newblock {\em arXiv preprint arXiv:2004.08955}, 2020.

\end{thebibliography}
}

\end{document}